\documentclass{article}
\usepackage{ijcai23}

\usepackage{times}
\usepackage{soul}
\usepackage{url}
\usepackage[hidelinks]{hyperref}
\usepackage[utf8]{inputenc}
\usepackage[small]{caption}
\usepackage{graphicx}
\usepackage{amsmath}
\usepackage{amsthm}
\usepackage{booktabs}
\usepackage{algorithm}
\usepackage{algorithmic}
\usepackage[switch]{lineno}

\title{Elastically-Constrained Meta-Learner for Federated Learning}

\author{
Peng Lan\footnotemark
\and
Donglai Chen\footnotemark[1]\and
Chong Xie\and
Keshu Chen\and
Jinyuan He\and \\
Juntao Zhang\and
Yonghong Chen\footnote{Contact Author}\And
Yan Xu
\affiliations
Linklogis Inc.\\
\emails
\{lanpeng, chendonglai, xiechong, chenkeshu, hejinyuan, zhangjuntao, chenyonghong, xuyan\}@linklogis.com,
}

\begin{document}

\maketitle
\footnotetext[1]{Equal contribution}
\begin{abstract}
    Federated learning is an approach to collaboratively training machine learning models for multiple parties that prohibit data sharing. One of the challenges in federated learning is non-IID data between clients, as a single model can not fit the data distribution for all clients.  Meta-learning, such as Per-FedAvg, is introduced to cope with the challenge. Meta-learning learns shared initial parameters for all clients. Each client employs gradient descent to adapt the initialization to local data distributions quickly to realize model personalization. However, due to non-convex loss function and randomness of sampling update, meta-learning approaches have unstable goals in local adaptation for the same client. This fluctuation in different adaptation directions hinders the convergence in meta-learning. To overcome this challenge, we use the historical local adapted model to restrict the direction of the inner loop and propose an elastic-constrained method. As a result, the current round inner loop keeps historical goals and adapts to better solutions. Experiments show our method boosts meta-learning convergence and improves personalization without additional calculation and communication. Our method achieved SOTA on all metrics in three public datasets.
\end{abstract}
\section{Introduction}
\label{sec:intro}

Federated learning (FL) is a privacy-preserving distributed machine learning algorithm that trains models by exchanging model information of all parties without exposure to their raw data \cite{yang2019federated}. Typical FL algorithms like Google's FedAvg \cite{mcmahan2017communication}, include a server that aggregates and distributes parameters as well as clients that own data. In each round of training, the server broadcasts the global model to clients. Clients use local datasets to train their models. After local training, the clients gain diverse models depending on the local distribution and capture latent information of the data. The server collects and aggregates clients' parameters to update the global model. Then the next round begins. 
Through the parameter aggregation, the message is passed across the clients iteratively over communication rounds. After training, the final aggregated model is distributed to clients for employment. In this way, information from clients is communicated to learn a better model than isolated training without data breach in the whole process.

Data heterogeneity is one of the challenges in FL. In real-world applications, clients' data are non-IID \cite{zhao2018federated}. FedAvg performs well in identical and independent distributed (IID) data but its performance suffers significantly on non-IID data, as one global model cannot fit different data distributions simultaneously. One solution is to allow personalized model per client for better local fitting, which is named personalized federated learning (PFL) \cite{kulkarni2020survey}.

Meta-learning-based models have been proposed to solve PFL. Meta-learning aims at training model initialization parameters (called meta-initialization) so that  they can adapt to any task quickly \cite{vanschoren2019meta,TimothyMHospedales2020MetaLearningIN}. In each training step, the meta-initialization $\varphi$ is updated on a batch of different tasks (called meta-batch). The tasks in meta-batch generate data from diverse distribution. The update for $\varphi$ can be viewed as a double-loop optimisation: inner and outer loop. The inner loop adapts meta-initialization $\varphi$ to each task via gradient descent on labeled data, in order to get task-specific solution. The outer loop updates meta-initialization based on the results of the outer loop. 

The procedure of meta-learning can be naturally modified to 
adapt to FL. By formulating each
client as a task, the per-task operations can be executed on each
client locally, and the aggregation operation for a meta-batch can be
done on the server.
FedMeta \cite{chen2018federated} and Per-FedAvg \cite{fallah2020personalized} introduce meta-learning and treat clients as different tasks. They search for the common optimal initial parameters for all tasks. Such parameters can quickly adapt to local data distribution by one or several steps of gradient update to fulfill the personalization goal.

\begin{figure}[ht]
    \includegraphics[scale=0.12]{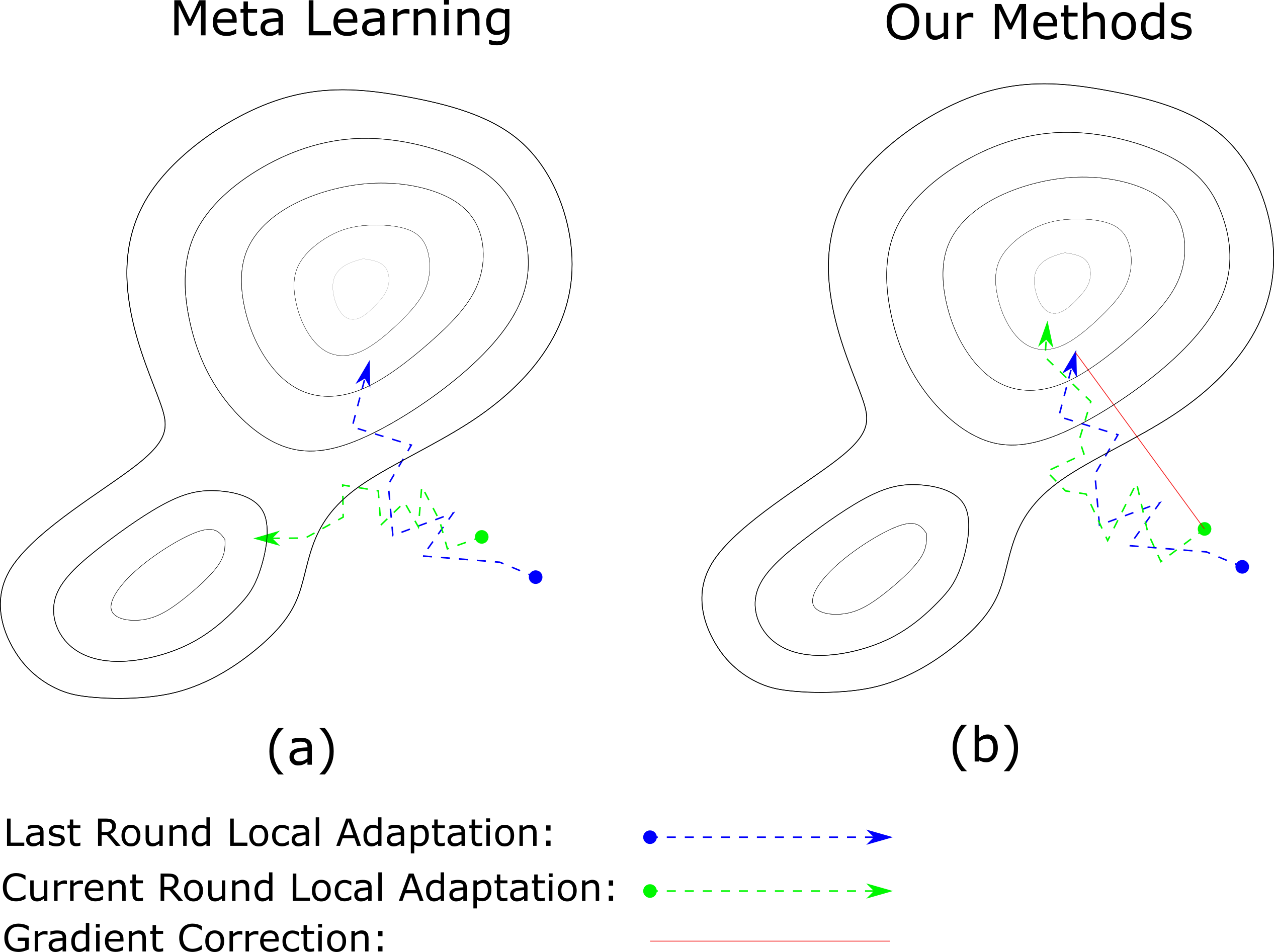}
    
    \caption{The meta-learning local adaptation tracks for the same task and different sampling time. The contour plots show levels of loss function at different places. Due to changes in meta-learning parameter training two sampling time and randomness of SGD, in (a) local adaptation may lean toward different local optima. while in (b) the update keeps stable goal with elastic constraint.}
    \label{Figure1}
    
\end{figure}

One limitation of meta-learning-based approaches is the fluctuation in meta-learning's training process. In the same client and different sampling loops, goals are unstable. That is caused by non-convexity of loss function and randomness of stochastic gradient descent (SGD). Specifically, in Figure \ref{Figure1}(a), in different sampling times, meta-learning's inner loop may lean towards different local optima, that is fluctuation in optimization goals.

\setlength{\intextsep}{0.5cm}

Whereas outer update relies on inner loop's fast adaptation results. For example in Reptile \cite{AlexNichol2018OnFM}, the results of inner loop are used as meta-initialization's update direction. Therefore, deviation of inner update would cause deviation of outer update and worsen the whole meta-learning optimization results.

To address the problem, we propose a method of constraint based on past update information. It aims to correct current inner loop optimization goal using historical local adaptation point in order to deal with outer update fluctuations. We discover that improper hard constraint, such as L2 would hinder the algorithm searching for better solutions. The algorithm gets stuck at suboptimal solutions and is slow to converge.

The purpose of the inner update is changed from fitting local data's ground truth to fitting the weighted average of historical models' prediction probability and ground truth. The goal takes historical models into account. When the historical model has a better fitting ability, the model updates towards a promising optimization direction and alleviates deviation caused by non-convexity and randomness. On the other hand, when the historical model fails to give useful information, the goal shifts to randomly smoothed ground truth. It motivates the model to find a better fit and prevents  suboptimal solutions. Our method incorporates both exploitation of historical fitting results and exploration of better solutions and improves meta-learning performance under federated scenarios.

We did extensive experiments on three public datasets. Results show that our algorithm  effectively alleviates fluctuation in meta-learning, accelerates convergence and achieves new SOTA.

We summarize our contributions as follows: 
\begin{itemize}
    \item We find meta-learning-based federated learning has fluctuation in loss through experiments and that using historical update information can alleviate the problem.
    \item Based on the first point, we propose a new implicit  regularization method. It uses historical information to constrain update direction and preserve the ability to search for global optimum. It accelerate meta-learning convergence and has better adaptation.
    \item Extensive experiments show our method achieves SOTA performance in three datasets and various statistical heterogeneity settings.
\end{itemize}

\section{RELATED WORK}
Our method is related to federated learning, meta-learning and regularization. The related works are three-fold. 
\subsection{Federated Learning}

Federated learning is 
a machine learning technique that preserve clients' privacy. Google's FedAvg \cite{mcmahan2017communication} is a basic one. It takes weighted average of clients' local trained model to get global model. Then it distributes to all clients as the initialize parameter for clients' local training. Those FL algorithms may not perform well if clients' data are non-IID. 

The following researches improve FedAvg for non-IID data. FedProx \cite{TianLi2018FederatedOI} adds constraint in the local training update. It uses the L2 distance between local parameter and global parameter as regularization term and make convergence process stable. SCAFFOLD \cite{SaiPraneethKarimireddy2020SCAFFOLDSC} proposed control variates to correct client-drift during local training and boost convergence of FL. In FedDyn \cite{acar2021federated}, risk objective for each device is dynamically updated, such that clients' models move towards global optimum.

The above methods use the same global model to adapt to different distributions in different clients. While some researches propose local model personalization solutions that can cope with the challenge more effectively. The model personalization can be realized in different ways. MOCHA \cite{smith2017federated} and FedEM  \cite{OthmaneMarfoq2021FederatedML} treat each client as a task and realized multi-task learning under federated scenarios. FedPer \cite{arivazhagan2019federated}, FedRep \cite{LiamCollins2021ExploitingSR} and LG-FedAvg \cite{PaulPuLiang2020ThinkLA} break down model to feature layers and classification layers. They use part of the model as local private layer to realize personalization. The FedPer and FedRep share feature layer and keep classification layer private, and the LG-FedAvg does the opposite. FedMatch \cite{JianguiChen2021FedMatchFL} further devises private patch for transformer structure to achieve personalized federated QA system. And PartialFed-Adaptive \cite{BenyuanSun2021PartialFedCP} automatically selects personalized layers for each client to realize cross-domain personalization. Clustered FL methods like CFL \cite{FelixSattler2019ClusteredFL} and IFCA \cite{AvishekGhosh2020AnEF} let clients in one cluster share a model to realize cluster-level personalization. Model mixture methods like APFL \cite{deng2020adaptive} and Fed-MGL  \cite{hanzely2020federated} train local models and global models simultaneously, and combine local models with global models to obtain final personalized model. FedAMP \cite{YutaoHuang2020PersonalizedCF} introduces an attentive message passing mechanism, which achieve personalized aggregate models by promoting parameter exchange between clients with similar data distribution. Similarly, KT-pFL \cite{JieZhang2021ParameterizedKT} proposes a trainable knowledge coefficient matrix, which reinforces the collaboration among similar clients and facilitate the communication using local soft predictions. The pFedHN \cite{AvivShamsian2021PersonalizedFL} brings hyper network into FL to generate personalized models. The pFedME \cite{CanhTDinh2020PersonalizedFL} introduces Moreau Envelopes to form a regularized loss, which leverages global model to achieve better local training.

Meta-learning-based FL approaches, such as FedMeta \cite{chen2018federated}, Per-Fedavg \cite{fallah2020personalized} are the most related ones to ours. 
They aim to learn better initial parameter for each client. They realize personalization in clients through one or several step gradient update to quickly adapt to local data distribution. FedMeta employed empirical analysis to validate the effectiveness of meta-learning under federated scenarios. Per-FedAvg theoretically proved its convergence rate. GPAL \cite{YounghyunPark2021FewRoundLF} also employs meta-learner for FL, but it handles the lack of data variability per client and limited communication budget, rather than PFL.

\subsection{Meta-Learning}
In meta-learning, models are trained by different tasks and learn the ability to quickly adapt to new tasks. One typical algorithm is LSTM Meta-Learner \cite{YutianChen2016LearningTL}. It uses training gradients in different time as the input of LSTM and treat model parameters as cell state, in order to exploit different tasks' commonality in gradient update and help optimization in new task parameters. MAML \cite{ChelseaFinn2017ModelAgnosticMF} aims to get an optimal initialization parameters through different training tasks, so that the model can adapt to new tasks by one or several steps of gradient descent. 

Past researches improve the meta-learning from different perspectives. iMAML \cite{AravindRajeswaran2019MetaLearningWI} introduces an implicit differentiation based method to reduce the computational and memory burden in MAML, and Reptile \cite{AlexNichol2018OnFM} proposes a first-order gradient-based meta-learner to further cut the computation overhead. PLATIPUS  \cite{ChelseaFinn2018ProbabilisticMM} and BMAML \cite{JaesikYoon2018BayesianMM} propose probabilistic or Bayes version of meta-learner, which generates the model distribution depending on the observed data. LEO \cite{AndreiRusu2018MetaLearningWL} achieves improved performances on few-shot regimes by optimizing the latent embedding of high-dimensional model parameter. MAML++ \cite{AntreasAntoniou2018HowTT} discusses the network architectures sensitivity of the meta-learning and makes various modiﬁcations to achieve better generalization, convergence and computational cost. UNICORN-MAML \cite{HanJiaYe2021HowTT} finds that meta learning needs more update steps in the inner loop and is sensitive to label permutation in meta test, and introduces a single-vector method for classification to achieve new SOTA in few-shot learning. Researches like ANIL \cite{AniruddhRaghu2020RapidLO} and BOIL \cite{JaehoonOh2020BOILTR} explore the how meta-learner works focusing on whether it achieve fast adaptation or feature reuse.

Meta-Learning is initially invented to solve few-shot learning and reinforcement learning. But its quick adaptation method has been widely used in other areas in the subsequent research. \cite{JaewoongShin2021LargeScaleMW} extends meta-learning to many-shot application. MLDG \cite{DaLi2017LearningTG} applies meta-learning in semantic parsing model, in order to learns a domain-agnostic model that can be applied to an unseen domain. FastDFKD \cite{fang2022up} uses meta-learning to speed up knowledge distillation. 

\subsection{Regularization}
Regularization is a common technique in machine learning to avoid overfitting in model training by reducing the magnitude of model parameters. Traditional regularization methods such as L1, L2 use the L1 and L2 norm as additional loss term to limit model complexity. 

EWC (Elastic Weight Consolidation, \cite{kirkpatrick2017overcoming}) is one of the regularization methods in continual learning to overcome catastrophic forgetting. It calculates models parameter importance weights. New model has larger constraint weight at high importance parameters. It helps keeping past important knowledge in continual learning and alleviate catastrophic forgetting. 

Knowledge distillation (KD) aims to transfer the prediction ability from one model (teacher) to another (student) \cite{gou2021knowledge}. Usually the teacher model has high prediction power. During KD, the student model learns the soft label from the teacher model, which may contain more information than the one-hot label. 
KD adds an additional loss term in the objective function. It can constrain student model's representation and keep part of teacher model's knowledge and realize goal of generalization. 

Knowledge distillation and label smoothing regularization (LSR) are related \cite{yuan2020revisiting}. LSR replaces one-hot labels with soft labels \cite{szegedy2016rethinking}. It can be viewed as KD with an uninformative teacher. KD can improve student model's performance in two scenarios. On one hand, the uninformative teacher can improve the student by the effect of label smoothing regularization in the KD process, on the other hand, the student can learn from the teacher if the teacher has high performance.

\section{Methodology}
\subsection{Problem Formulation}
The objective of multi-party machine learning is to maximize the overall performance of the models for all parties, which is formally
\begin{equation}
    \mathop{min}_{(\theta_{1},\ldots,\theta_{N})} \frac{1}{N} \sum_{i=1}^{N}
l_{i}(\theta_{i}),
\end{equation}
where $N$ is the number of clients, $\theta_{i}: \mathop{R}^{d} \rightarrow Y$ is the learning model for \emph{i}-th client. The $l_{i}$ is expected loss over local distribution. Namely $l_{i}:=E_{(x,y)\sim D_{i}}l(\theta_{i}(x),y)$, where $x \in \mathop{R}^{d}$ and $y \in Y$ are the features and label for a data point,
$D_{i}$ is the data distribution for \emph{i}-th client and $l$ is the loss function such as cross entropy.

To outperform isolated training, one straightforward option is centralized training, in which the server collects data 
from all parties thus enriching the overall training data. However, this is not practical, as participants with privacy and security concerns usually prohibit any data breach. In FL, the server communicates model parameters rather than data with clients so that data are isolated and the performance is improved. 

Standard FL algorithms like FedAvg aggregate the passed messages and get one global model for all clients, formally $\theta=\theta_{1}=\ldots=\theta_{N}$. In IID setting which assumes $D_{global}=D_{1}=\ldots=D_{N}$, one model that fits one distribution $D_{global}$ is able to generalize on every client. The non-IID setting brings significant performances degradation, as one model $\theta$ is hard to fit $N$ diverse distributions $D_{1} \neq \ldots \neq D_{N}$. Personalized federated learning (PFL) solves this more challenging circumstance by diversifying models $\theta_{1} \neq \ldots \neq \theta_{N}$ to fit each local distribution.

\subsection{Meta-learner for Personalized Federated Learning}

Typical meta learning methods like MAML and Reptile \cite{AlexNichol2018OnFM}, achieve learn-to-learn via training an initialization $\varphi$ (often named meta-initialization) which can quickly adapt to any task. In every training step, the meta-initialization $\varphi$ is updated on a batch of different tasks (often called meta-batch). The tasks in meta-batch generate data from diverse distribution ${D_{1} \neq D_{2} \neq \ldots}$. The update for $\varphi$ can be viewed as a double-loop optimisation: inner and outer loop. The inner loop adapts meta-initialization $\varphi$ to each task via gradient descent on labeled data, in order to get task-specific solution ${\theta_{1}, \theta_{2}, \ldots}$. The outer loop updates meta-initialization based on ${\theta_{1}, \theta_{2}, \ldots}$ with different methods depending on the variant of meta-learners.
 
MAML is a typical meta-learner. It divides the task data as support set and query set for the two loops respectively. The motivation of this division is using a separate data set to evaluate the generalization performance of the local adaptation on a task. In the inner loop for $i$-th task, task-specific solution $\theta_{i}$ is got via one gradient descent step from $\varphi$ on support set, which is formally
$\theta_{i} \leftarrow \varphi - \varepsilon_{in} \nabla_{\varphi} l_{i}(\varphi)$.

 In the outer loop, based on the loss of $\theta_{i}$ on query set, $\varphi$ is updated via another gradient descent, 
\begin{equation}
\varphi \leftarrow \varphi - \varepsilon_{out} \sum_{i} \nabla_{\varphi}  l_{i}(\theta_{i}).
\end{equation}
 The $\varepsilon_{in}$ and  $\varepsilon_{out}$ are the learning rates for inner and outer loops respectively, and the losses on support and query sets for $i$-th task are both denoted as $l_{i}$ for notation simplicity.
 As $\theta_{i}$ consists of the derivative with respect to $\varphi$, the outer update of MAML involves the second-order derivative operation, which is computationally intensive. Reptile is a ﬁrst-order gradient-based meta-learner proposed by OpenAI \cite{AlexNichol2018OnFM}. It removes the second-order derivative operation in outer loop while achieves competitive performance as MAML. The following demonstrates Reptile double-loop optimisation: \\
\textbf{inner update}: \\
initialize $\theta_{i} \leftarrow \varphi$, and do
\begin{equation}
\theta_{i} \leftarrow \varepsilon_{in} \nabla_{\theta_{i}} l_{i}(\theta_{i})
\end{equation}
for $\tau$ steps, \\
\textbf{outer update}: \\
\begin{equation}
\varphi \leftarrow \varepsilon_{out} \sum_{i}(\theta_{i}-\varphi).
\end{equation}
Inspired by Reptile, our method employs similar outer loop strategy for computational overhead reduction.

We define the notations used in previous meta-learners for FL, like Per-FedAvg and FedMeta. First of all, the server initializes the meta-parameter $\varphi^{0}$ and starts up communication with clients. The communication iterates $T$ rounds. At \emph{t-th} communication round, a clients subset $I^{t}$ is sampled for training. The server sends last round meta-initialization $\varphi^{t-1}$ to all clients sampled. Each client $i$ does inner update from $\varphi^{t-1}$ to get $\theta^{t-1}_{i}$, and then calculate the per-task outer update component $\Delta \varphi_{i}^{t-1}$ based on $\theta^{t-1}_{i}$. These per-task components are sent back, and the server completes the aggregation operation of outer update:
\begin{equation}
\varphi^{t} \leftarrow \varphi^{t-1} + \varepsilon_{out} \frac{1}{\vert I^{t} \vert} \sum_{i \in I^{t}} \Delta \varphi_{i}^{t-1}.
\end{equation}
 Then the next round starts. After training, each client fine-tunes from the meta-initialization $\varphi^{T}$ on the local data to get task-specific solution $\theta_{i}$ for final employment. 

\begin{table}
\begin{tabular}{l}
\hline \textbf{Algorithm 1 FedEC} \\
\hline \textbf{Parameters}: \\
number of clients $N$; \\
sampling rate $r$; \\
number of communication rounds $T$; \\
outer loop learning rate  $\varepsilon_{out}$; \\
inner loop steps $\tau$; \\
inner loop learning rate $\varepsilon_{in}$; \\
regularization weight $\alpha$; \\
\\
\textbf{Overall Algorithm:} \\
1:  initialize $\varphi_{0}$ at random\\
2:  \textbf{for} $t = 1, 2, \cdots, T$ \textbf{do}\\
3:  \quad sample clients set $I^{t}$ of size $rN$ \\
4:  \quad \textbf{for} each client $i \in I^{t}$ in parallel \textbf{do}\\
5: \qquad  $\theta^{t,\tau}_{i} \leftarrow \boldsymbol{InnerUpdate_{i}}(\varphi^{t-1})$ \\
6: \qquad  client $i$ sends updated parameter $\theta^{t,\tau}_{i}$ to server \\
8: \quad\textbf{end for} \\
9: $\boldsymbol{OuterUpdate}$: \\
\quad \quad  $\varphi^{t} \leftarrow \varphi^{t-1} + \varepsilon_{out}\frac{1}{rN} \sum_{i \in I^{t}} (\theta^{t,\tau}_{i} - \varphi^{t-1})$ \\
10: \textbf{end for} \\
\\
$\boldsymbol{InnerUpdate_{i}}$: \\
\textbf{input} $\varphi^{t-1}$ \\
11: $\theta^{t,0}_{i}\leftarrow \varphi^{t-1}$\\
12: \textbf{for} $s = 1, 2, \cdots, \tau$ \textbf{do}\\
13: \quad \textbf{if} $\hat{\theta_{i}}$ initialized \textbf{do} \\
14: \qquad $\theta^{t,s}_{i} \leftarrow \theta^{t,s-1}_{i} - \varepsilon_{in} \nabla l_{i}(\theta^{t,s-1})$ \\
15: \quad \textbf{else} \\
16: \qquad $\theta^{t,s}_{i} \leftarrow \theta^{t,s-1}_{i}-\varepsilon_{in}\nabla(l_{i}(\theta^{t,s-1}_{i})+$ \\ \qquad \qquad $\alpha c_{i}(\theta^{t,s-1}_{i},\hat{\theta_{i}})) $ \\
17: \textbf{end for} \\
18: initialize or update local memory module $\hat{\theta_{i}} \leftarrow \theta^{t,\tau}_{i}$ \\
\textbf{return} $\theta^{t,\tau}_{i}$
\\
\hline
\end{tabular}
\end{table}

\subsection{Model Overview}
\label{model}
As stated in the introduction, inner loop fluctuation worsen the performance of meta-learner for FL. To cope with this problem, we proposed a elastically-constrained meta-learner for Federated Learning (abbreviated as FedEC). FedEC leverages the history optimisation information to facilitate efficient and stable update in the inner loop. We also introduce first-order gradient-based update strategy in the outer loop, which reduces the computation overhead. Our method is summarized in Algorithm 1. The detailed procedures of FedEC are described as follows.

For a communication round $t$ between $1$ and $T$, the server samples a subset of clients $I^{t}$. The clients have heterogeneous distributions, so every client is treated as a task and the $I^{t}$ corresponds to a meta-batch. The server sends the latest meta-initialization $\varphi_{t-1}$ (for first communication round $\varphi_{0}$ is randomly initialized) to all sampled clients. Then the inner update is executed on clients in parallel following the standard FL. However, in our method, inner update depends on not only current meta initialization $\varphi_{t-1}$ but also the historical model $\hat{\theta_{i}}$ by introducing the constraint. The $\hat{\theta_{i}}$ stores the last inner update result, and it is updated every time the $i$-th client sampled. Thus, for most rounds the inner loss is changed from $l_{i}(\theta)$ to $l_{i}(\theta) + \alpha c_{i}(\theta,\hat{\theta_{i}})$, where $c_{i}(\theta,\hat{\theta_{i}})$ refers to the constraint and $\alpha$ refers to its weight. When the $i$-th client is sampled for the first time, the constraint term is set to 0 as $\hat{\theta_{i}}$ is not initialized by any training. Initialized from $\varphi_{t-1}$, the parameter trains on the local data set for $\tau$ epochs. Formally the constrained inner loop loss $\tilde{l}_{i}$ for $i$-th client is
\begin{equation}
\tilde{l}_{i}(\theta) = \begin{cases}
l_{i}(\theta) + \alpha c_{i}(\theta, \hat{\theta_{i}}) & \hat{\theta_{i}}\quad initialized \\
l_{i}(\theta) & else. \\
\end{cases}
\end{equation}

The training can use any gradient methods such SGD, RMSprop or Adam. The result of $i$-th client at $t$-th communication round after $\tau$ epochs of local adaptation is denoted as $\theta^{t,\tau}_{i}$, which is tested as employed model at current communication round. The $\theta^{t,\tau}_{i}$ is then copied and sent to server for outer update, and also stored into $\hat{\theta_{i}}$ to constrain next round inner loop. The inner loop is formulated in the $InnerUpdate_{i}$ part of Algorithm 1.

FedEC adopts a first-order approximation strategy for outer update. Concretely, given all local adaptations $\theta^{t,\tau}_{i}$ for $i$ from $1$ to $N$, FedEC approximates the per-task component of outer update by $\theta^{t,\tau}_{i}-\varphi^{t-1}$. We denote $U_{i}^{\tau}$ as local update on data from $i$-th client for $\tau$ epochs, and $\theta^{t,\tau}_{i}$ can be written as
$\theta^{t,\tau}_{i}=U_{i}^{\tau}(\varphi^{t-1})$.

So the outer update component for $i$-th client is
\begin{equation}
\Delta \varphi_{i}^{t-1} = U_{i}^{\tau}(\varphi^{t-1})-\varphi^{t-1}, 
\end{equation}
which is the direction from meta-initialization to the local adaptation. The outer update takes the average for components of all clients, formally
\begin{equation}
\varphi^{t} \leftarrow \varphi^{t-1} + \varepsilon_{out} \frac{1}{\vert I^{t} \vert} \sum_{i \in I^{t}} \Delta \varphi_{i}^{t-1}.
\end{equation}
This averaging aggregation acts like a voting process, and the averaged direction drives meta-initialization to a point that is good for most local adaptation. Moreover, FedEC significantly reduces the communication overhead and simplifies the procedure, as the second-order derivative and data-split operations are removed.

The details of the elastic constraint is illustrated below. The constraint term is 
\begin{equation}
c_{i}(\theta,\hat{\theta_{i}}) = KL(\hat{\theta_{i}}(x), \theta(x)),
\end{equation}
where $D_{KL}$ is the Kullback-Leibler divergence and $\hat{\theta_{i}}$ is the reserved model. 
Considering the local empirical loss 
\begin{equation}
l_{i}(\theta) = CE(y, \theta(x)),
\end{equation}
where $CE$ is the cross entropy, $\theta(x)$ is the predicted probabilities of training model and $y$ the one-hot embedding of the ground truth. The constrained loss can be written as
\begin{equation}
    l_{i}(\theta) + \alpha c_{i}(\theta,\hat{\theta_{i}}) = CE(y, \theta(x)) + \alpha KL(\hat{\theta_{i}}(x), \theta(x))
\label{eq}
\end{equation}


We introduce a generated distribution $q(x)$ as
\begin{equation}
q(x)=\frac{1}{1+\alpha}y+\frac{\alpha}{1+\alpha}\hat{\theta_{i}}(x).
\end{equation}
The $q(x)$ can be seen as a distribution generated from weighted average of one-hot embedding and $\hat{\theta_{i}}$ predictions.
Substituting $q(x)$ into the equation (\ref{eq}), (\ref{eq}) equals to
\begin{equation}
    (1+\alpha)CE(q(x), \theta(x))+\alpha \hat{\theta_{i}}(x) \log(\hat{\theta_{i}}(x)).
\end{equation}


where the second term $\alpha \hat{\theta_{i}}(x) \log(\hat{\theta_{i}}(x))$ is a constant. The constrained loss changes the optimization target by combining the one-hot embedding and the distribution of previous prediction. Compared with the unconstrained loss $l_{i}(\theta)$, this combination encourages $\theta(x)$ not only to predict higher probability corresponding to true label, but also to imitate the prediction distribution for historical model. There exists two circumstances for historical model $\hat{\theta_{i}}$ in the training. One is that when historical model does not fit the local distribution properly, which often happens in the early communication. Thus, the prediction of $\hat{\theta_{i}}$ can be seen as a random distribution. This plays a role of regularization by smoothing target of training. The training target is one-hot embedding of ground truth in unconstrained loss, and the model learns to predict more extreme distribution, thus prone to over-fitting. While the constrained target reduces the predicted probability gap between classes, so the generalization of model is kept. The other is that $\hat{\theta_{i}}$ gains fitting ability after certain rounds of learning.  Thus the introducing of historical prediction facilitates exploitation of previous knowledge, thus the update can avoid local optimal and converge quickly to equivalent performance of historical adaptation. While the ground truth part keep model exploring better fitting for training data. So the model would not be stuck at a historical solution and the continual exploration for new solution is kept.

\section{EXPERIMENT}
We do extensive experiments to evaluate FedEC's performance by  comparing with different benchmarks, as well as L2 regularization under different experiment settings. In the case study, visualizations show the effectiveness of FedEC to deal with model fluctuation.

\subsection{Dataset}\label{Dataset}
To evaluate the performance of our method, we use three public datasets to build benchmark.
CIFAR10 is a dataset of 60000 images with sizes of 32*32 from 10 classes. Classes include car, plane, bird, etc. CIFAR100 is similar to CIFAR10 with 60000 images. Its difference is that this dataset further classifies images into 20 super classes and 100 subclasses. FEMNIST \cite{SebastianCaldas2018LEAFAB} is built by partitioning the data in Extended MNIST.
Data summary statistics are shown in Table \ref{Dataset information}.
\setlength{\intextsep}{0.1cm}
\begin{table}[htb]
    \renewcommand\arraystretch{1.0}
    
    \centering
    \scalebox{1}
    {
    \setlength{\tabcolsep}{7.0 pt}  
    \begin{tabular}{c c c c c c}
        \hline 
        \textbf{DataSet} & \textbf{total} &
        \textbf{train} &  \textbf{test} & 
        \textbf{classes}
        \\
        \hline
        CIFAR10 & 60000 & 50000 & 10000 & 10  \\
        CIFAR100 & 60000 & 50000 & 10000 & 100 \\
        FEMNIST & 34674 & 31123 & 3551 & 10  \\ 
        \hline 
    \end{tabular}
     
    }
    \caption{\centering Details of three Datasets}
    \label{Dataset information}
\end{table}

\subsection{Experiment Setting}\label{4.2 experiment setting}

\subsubsection{Model structure}
We evaluate on different model structures. For CIFAR10 and CIFAR100, the main model structure are two convolutional layers and three fully connected layers. For FEMNIST, the structure is three fully connected layers. All models use ReLU as the activate function. 
For evaluation metric, we use the average \emph{acc} values of the local test set of all clients when parameter aggregation has been completed for each communication round.

\subsubsection{Data splitting method} 
We split the data into training set and test set in a non-IID way. For CIFAR10 and CIFAR100, we set classes at each client to control the level of non-IID. In this scenario, each client has the same number of classes and different combination of classes. Number of images of each class at a client equals to the total number of images at that class divided by the number of clients. For instance, assume we set 100 clients in total and each client contains two distinct classes, and CIFAR10 is chosen to conduct the experiment. The data splitting method is described as follows: initially, separate each class in CIFAR10 into 20 pieces (where 20 is computed via $2\times100\div10$); then, randomly allocate each client with two classes (duplication is permitted), the total number of allocation for each class is identical; ultimately, specific data in the corresponding class is assigned to each client.  For each communication round, we randomly selects 10\% of clients to participate in federated training. Additionally, the allocated data to each client is different. By changing the number of classes per client, the level of statistical heterogeneity is tuned. Specifically, lower number classes per client, which means fewer shared classes between clients, corresponds to a more heterogeneous case. We also experiment on various number of clients, the detailed statistical result of experiments on different datasets is in Table \ref{benchmark}. The settings are consistent with \cite{LiamCollins2021ExploitingSR}.
To examine the strength of the model-agnostic methods, we implement various common networks on different datasets.

\subsubsection{Hardware resource} We use one Nvidia A100 card to test our model. A100 has 40G video memory. CPU is Intel(R) Xeon(R) Gold 6248R @ 3.0Ghz.

    
    

\setlength{\intextsep}{0.1cm}
\begin{table*}[htb!]
    \renewcommand\arraystretch{1.0}
     
    \centering
    \scalebox{1}
    {
    \setlength{\tabcolsep}{8.0 pt}  
    \begin{tabular}{c c c c c c c c}
        \hline 
        \textbf{Model}
        & \multicolumn{3}{c}{\textbf{CIFAR10}} & \multicolumn{2}{c}{\textbf{CIFAR100}} & \multicolumn{1}{c}{\textbf{FEMNIST}} \\
        \cline{2-8}
         & $\mathbf{(100,2)}$ & $\mathbf{(100,5)}$ & 
         $\mathbf{(1000,2)}$ & $\mathbf{(100,5)}$ & $\mathbf{(100,20)}$ & $\mathbf{(150,3)}$ \\
        \hline
        Local Only & 89.79 & 70.68 & 78.30 & 75.29 & 41.29 & 40.12 \\
        \hline
        FedAvg & 42.65 & 51.78 & 44.31 & 23.94 & 31.97 & 31.26 \\
		SCAFFOLD & 37.72 & 47.33 & 33.79 & 20.32 & 22.52 & 17.65 \\
		pFedME & 60.23 & 70.01 & 62.33 & 23.64 & 34.66 & 47.66 \\
		LG-FedAvg & 84.14 & 63.02 & 77.48 & 72.44 & 38.76 & 41.38 \\
		L2GD & 81.04 & 59.98 & 71.96 & 72.13 & 42.84 & 46.56 \\
		APFL & 83.77 & 72.29 & 82.39 & 78.20 & 55.44 & 52.46 \\
		FedPer & 87.13 & 73.84 & 81.73 & 76.00 & 55.68 & 59.07 \\
		Per-FedAvg & 82.27 & 67.2 & 67.36 & 72.05 & 52.49 & 71.51 & \\
        FedRep & 87.70 & 75.68 & 83.27 & 79.15 & 56.10 & 59.21 \\  
        FedEC-wo  & 91.03 & 82.26 & 90.87 & 80.29 & 58.07 & 53.27 & \\
        FedEC-l2 & 87.38 & 71.15 & 87.13 & 74.90 & 43.09 & 72.03 &  \\
        \hline
        FedEC & \textbf{92.35} & \textbf{83.78} & \textbf{92.11} & \textbf{81.75} & \textbf{59.1} & \textbf{77.3} \\ 
        \hline 
    \end{tabular}
    
    }
    \caption{\centering  Average accuracy of local test over the ﬁnal 10 rounds out of T = 100 total rounds, where $(x, y)$ denotes  $x$ clients and $y$ classes per client}
    \label{benchmark}
\end{table*}

\subsection{Benchmark}\label{4.3 Benchmark}

We compare FedEC with other baseline methods thoroughly. Isolated training is also added to compare, which is denoted as local only. In Table \ref{benchmark}, we show the average acc at the last ten rounds for each method under six groups of parameters. FedEC outperforms the current best model in almost all criteria. We find that traditional FL methods such as FedAvg, SCAFFOLD \cite{SaiPraneethKarimireddy2020SCAFFOLDSC} fall behind the other personalized FL methods. LG-FedAvg \cite{PaulPuLiang2020ThinkLA}, L2GD \cite{FilipHanzely2021FederatedLO} perform worse than local only. APFL \cite{deng2020adaptive}, FedPer \cite{arivazhagan2019federated} and FedRep \cite{LiamCollins2021ExploitingSR} improve from previous models, but they are lower than FedEC. As the data heterogeneity increases, such as the more number of classes in local clients or more clients, the gap between FedEC and other methods are more significant. This shows that our model can effectively tackle non-IID problem and outperform other methods including traditional meta-learning-based methods.

\subsection{Ablation study for the constraint}\label{4.4 ablation study}
We conduct ablation study and compare results of L2 regularization. The results are in Table \ref{benchmark}. We compare our method, FedEC with traditional meta-learning-based FL models (called FedEC-wo).

Methods with elastic regularization improve the acc for about 1-2 percents on CIFAR10 and CIFAR100 and 24 percents on FEMINST, but L2 regularization method has lower metrics in three datasets. The more significant gap happens on FEMINST is reasonable, as FEMINST, with more complicated images and smaller smaller data size, corresponds to a more challenging setting.
The possible reason for failure of L2 is that it poses a more rigid regularization compared to our method. The goal of regularization is to let current model adapt faster to equivalent optimum as historical model. The L2 imposes hard constraint on the parameter by only allowing parameter searching within a hypersphere with historical model as the center. This restricts the expressiveness of the model severely and worsen model performance. While elastic constraint allows any parameter searching direction as long as current model tracks the historical prediction distribution, which allows the model to search for better solutions.  


\setlength{\intextsep}{0.8cm}
\begin{figure}[ht!]
    \includegraphics[scale=0.5]{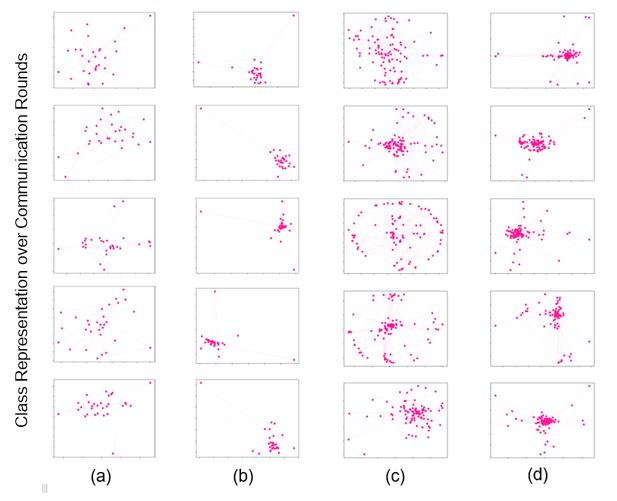}
    
    \caption{The t-SNE visualization for embedding changes over training rounds: (a) and (c) corresponds to FedEC-wo over 30 rounds and (b) and (d) corresponds to FedEC-wo over 100 rounds.}
    \label{visualization}
    
\end{figure}

\subsection{Case Study}\label{4.5 visualization}

We compare the change of local data embedding of FedEC-wo and FedEC in different communication rounds. We record multiple clients' average local feature representation after update at each communication round, and display their changes in consecutive multiple rounds using t-SNE \cite{van2008visualizing}. As shown in Figure \ref{visualization} (a) and (b) represent changes of embeddings of FedEC-wo and FedEC in the first 30 rounds. (c) and (d) show their pattern change in the first 100 rounds. Each row shows the change of one client in different rounds and different methods. Each dot represents the reduced dimension feature representation in one round.

From the comparison of the pair of (a) and (b) and the pair of (c) and (d), in the same iteration round, FedEC-wo has large change in the feature, sparse distribution of the reduced dimension feature representation and cannot effective converge due to model update fluctuation. FedEC can effectively solve this problem. Its features has small changes and has centralized distribution.

We compare (a) and (c) and find that FedEC-wo has multiple clusters. Its models jump between different clusters in the update process. This phenomenon validates our hypothesis in the introduction. Local models have multiple convergence points. Due to the randomness of sampling, the initialization of model parameters are much different from last round sampling local update. The update in different round converges to different optima. The inconsistency of optimization direction cause fluctuation in the local model update. We can see from (b) and (d), FedEC can effectively solve this problem. As the communication goes on, FedEC has gradual change apart from isolated points. There is only one cluster. FedEC effective alleviate model fluctuation.

\section{CONCLUSION}

In this work, we propose FedEC to deal with optimization fluctuation in traditional meta-learning-based FL models. FedEC incorporates Reptile in FL and learns from past models. It is immune to the discrepancy of goals in the same client in different sampling rounds. 
We compare our method with baseline methods in three public datasets under different heterogeneity conditions. The results show that our method effectively alleviates model fluctuation and improve personalization ability for client models. In the future, we will explore data heterogeneity in FL and expand FedEC to other personalized FL methods.

\bibliographystyle{named}
\bibliography{ijcai23}

\end{document}